%% file: main.tex
\documentclass[10pt, a4paper]{article}

\usepackage{lrec-coling2024}

\usepackage{booktabs}
\usepackage{multirow}
\usepackage{multicol}

\usepackage{caption}
\usepackage{subcaption}

\usepackage{amsmath}

\definecolor{seaborn-blue}{HTML}{4C72B0}
\usepackage{tcolorbox}
\tcbset{on line,left=0pt,right=0pt,top=0pt,bottom=0pt,colframe=white,colback=seaborn-blue!30}

\usepackage{enumitem}
\usepackage{amssymb}
\usepackage{pifont}
\newcommand{\cmark}{\ding{51}}
\newcommand{\xmark}{\ding{55}}
\newlist{mcanswers}{itemize}{1}
\setlist[mcanswers]{label=$\square$,topsep=0pt,itemsep=-1ex,leftmargin=3ex}

\title{Automatic Generation and Evaluation of Reading Comprehension Test Items with Large Language Models}

\name{Andreas Säuberli, Simon Clematide} 

\address{Department of Computational Linguistics, University of Zurich \\
         \{andreas, simon.clematide\}@cl.uzh.ch\\}

\abstract{
  Reading comprehension tests are used in a variety of applications, reaching from education to assessing the comprehensibility of simplified texts. However, creating such tests manually and ensuring their quality is difficult and time-consuming. In this paper, we explore how large language models (LLMs) can be used to generate and evaluate multiple-choice reading comprehension items. To this end, we compiled a dataset of German reading comprehension items and developed a new protocol for human and automatic evaluation, including a metric we call \emph{text informativity}, which is based on guessability and answerability. We then used this protocol and the dataset to evaluate the quality of items generated by Llama 2 and GPT-4. Our results suggest that both models are capable of generating items of acceptable quality in a zero-shot setting, but GPT-4 clearly outperforms Llama 2. We also show that LLMs can be used for automatic evaluation by eliciting item reponses from them. In this scenario, evaluation results with GPT-4 were the most similar to human annotators. Overall, zero-shot generation with LLMs is a promising approach for generating and evaluating reading comprehension test items, in particular for languages without large amounts of available data.
 \\ \newline  }

\begin{document}

\maketitleabstract

\section{Introduction}

Assessing reading comprehension is not only a crucial part of language testing in an educational context, it is also useful in many scenarios related to evaluation in natural language processing (NLP) -- for example, when evaluating the comprehensibility of automatically simplified texts \citep{Alonzo2021,Leroy2022,Saeuberli2024}, benchmarking the natural language understanding capabilities of large language models (LLMs) \citep{Lai2017,Bandarkar2023}, or determining factual consistency in text summarization \citep{Wang2020a,Manakul2023}. Multiple-choice tests are the most common way of assessing human reading comprehension because administering and grading them is simple. However, designing good multiple-choice reading comprehension (MCRC) items which actually test comprehension (as opposed to other things like the test taker's world knowledge or the readability of the item itself) is notoriously difficult \citep{Jones2020,Jeon2020}. Given the recent advancements in the zero-shot capabilities of LLMs \citep{Wei2021,Ouyang2022}, automatically generating MCRC items appears to be a promising option.

Evaluating MCRC items poses an additional challenge. While test developers in language assessment rely on extensive expert reviews and large pilot studies to determine the quality of test items \citep{Green2020,Gierl2021}, these evaluation methods are not practicable for fast-paced and iterative development of NLP models. In NLP research, there is still no consensus on evaluation methodologies and a lack of valid metrics for automatic evaluation \citep{Circi2023,Mulla2023}.

\begin{figure}
  \includegraphics[width=\columnwidth]{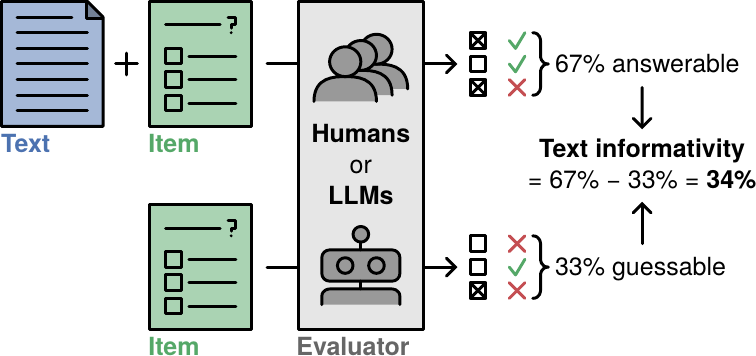}
  \caption{Our evaluation protocol measures the answerability and guessability of MCRC items by letting high-performing humans or LLMs respond to them with and without seeing the text. The text informativity metric is the difference between answerability and guessability and denotes to what degree the text informs the item responses.}
  \label{fig:protocol}
\end{figure}

In this paper, we address both the generation and the evaluation of MCRC items in the German language. We propose a new evaluation metric called \textbf{text informativity} combining answerability and guessability and use it for human and automatic evaluation. Our main contributions can be summarized as follows:

\begin{enumerate}
  \item We compile a dataset of German MCRC items from online language courses.
  \item We present a protocol for human evaluation of MCRC items and use it to evaluate items generated by two state-of-the-art LLMs.
  \item We demonstrate that the same protocol can also be used for automatic evaluation by replacing the human annotators with LLMs.
\end{enumerate}

\section{Background and Related Work}

\subsection{Automatic Item Generation}

Automatic item generation (AIG) has been of interest in educational and psychological assessment for several decades \citep{Haladyna2013}. Until now, rule-based approaches based on manually written templates have been used in these fields \citep{Lai2013,Circi2023}. Recent NLP research introduced neural approaches and especially pre-trained transformer models to generate comprehension questions \citep{Yuan2017,Du2017,Zhou2017,Gao2019,Lopez2020,Berger2022,Rathod2022,Ghanem2022,Uto2023,Fung2023}, multiple-choice distractors \citep{Maurya2020,Shuai2021,Xie2022}, or entire MCRC items based on a text in an end-to-end fashion \citep{Jia2020,Dijkstra2022}. Several works have reported promising results using zero-shot or few-shot prompting of LLMs \citep{Attali2022,Raina2022a,Kalpakchi2023}. While most previous research has focused on the English language, our work is the first to evaluate LLMs for zero-shot generation of German reading comprehension items.

\subsection{Evaluation of Generated Items}

Most NLP works on question generation and AIG report reference-based similarity metrics borrowed from machine translation or text summarization, such as BLEU, ROUGE, and METEOR \citep{Amidei2018,Circi2023,Mulla2023}. These metrics are unsuitable for generating MCRC items because similarity does not imply high quality for this task. Human evaluation is mostly done by asking experts or crowd workers to rate generated items in terms of fluency, relevance, difficulty, and other categories \citep[e.g.][]{Jia2020,Gao2019,Ghanem2022,Uto2023}. \citet{Attali2022} is a notable exception, conducting both expert reviews and a large-scale pilot study to evaluate LLM-generated test items.

Several studies have examined the possibility of using question answering (QA) models to evaluate generated items instead of human test takers. Most commonly, this is done by letting a QA model respond to the items and equating a high response accuracy to good answerability \citep{Yuan2017,Klein2019,Shuai2021,Rathod2022,Raina2022a,Uto2023}. In addition to answerability, \citet{Berzak2020}, \citet{Liusie2023}, and \citet{Raina2023} also measured guessability by benchmarking the model's ability to answer the items without seeing the text. Finally, \citet{Lalor2019} and \citet{Byrd2022} used large ensembles of models responding to human-written items and applied item response theory to determine psychometric measures such as difficulty and discrimination. Our evaluation protocol builds on these ideas and additionally leverages the recent advances in the natural language understanding capabilities to simplify and improve automatic evaluation.

\section{Evaluation Protocol}
\label{sec:protocol}

We propose a protocol for evaluating MCRC test items, including a new metric we call \textbf{text informativity} for evaluating an item's capability of measuring reading comprehension. It involves measuring the response accuracy of high-performing test takers when they have access to the text (\textbf{answerability}) and comparing it to their response accuracy when guessing the correct answer without seeing the text (\textbf{guessability}). To obtain these accuracies from human test takers, we first show them the items without the corresponding text and ask them to guess the correct answers. We then reveal the text and let them answer the same items again. Text informativity is then calculated as the difference between answerability and guessability. Intuitively, this metric represents to what degree the information extracted from the text helps the test takers to answer the test items. Since reading comprehension is essentially the ability to extract meaningful information from a text, a high text informativity indicates that the item actually measures the comprehension of the given text.

To apply this protocol for automatic evaluation, we replace human test takers with LLMs and we design prompts to elicit item responses twice for each item; once the text is included in the prompt, and once the model is instructed to guess the correct answers based on world knowledge. The assumption (which we are going to test) is that the LLMs are comparable to highly proficient human readers in terms of world knowledge and comparable reading comprehension capabilities.

Figure \ref{fig:protocol} illustrates the evaluation protocol. In the experiments described below, we will apply it to human-written and automatically generated items and compare the results to subjective ratings of item quality.

\section{Experimental Setup}

\subsection{Data}

We compiled a dataset of German texts and MCRC items from free online language courses\footnote{\url{https://learngerman.dw.com}} offered by \emph{Deutsche Welle} (DW), a broadcast company based in Germany. The target users for these courses are non-native speakers. We included the lessons from the \emph{Top-Thema} course\footnote{\url{https://learngerman.dw.com/de/top-thema/s-55861562}}, which consists of news articles which were summarized and simplified to match the B1 level in the Common European Framework of Reference for Languages (CEFR). The average text length is 327 tokens (\emph{spaCy} tokenization). Each text comes with several types of exercises, including three MCRC items. Almost all of these have three answer options, and in 66\% of the items, the user is allowed to select multiple answer options as correct. For simplicity, we will treat all items as if multiple correct answer options were possible.

We randomly selected 50 texts and all corresponding MCRC items as a test set for the experiment. For the human evaluation, we only used a subset of ten texts to reduce the workload for the annotators.

Scripts for scraping and preprocessing the dataset are available on GitHub\footnote{\url{https://github.com/saeub/dwlg}}. The dataset itself is currently not licensed for redistribution. We hope to publish the dataset for research purposes in the near future to enable more reproducible research.

\subsection{Models}

We selected two state-of-the-art instruction-tuned LLMs for generating MCRC items and as evaluators for the automatic evaluation:

\begin{enumerate}
  \item Llama 2 Chat (70B parameters; \texttt{meta-llama/Llama-2-70b-chat-hf} on Hugging Face) \citep{Touvron2023}
  \item GPT-4 (unknown model size; snapshot \texttt{gpt-4-0613}) \citep{OpenAI2023}
\end{enumerate}

\subsection{Zero-Shot Item Generation}

For each of the 50 texts, we prompted the two LLMs to generate three MCRC items with three answer options each, including which answer options were correct (refer to Appendix \ref{app:prompts} for the full prompts). We used a sampling temperature of 0 (i.e., greedy decoding) for both models.

Since Llama 2 is an English-centric model, it sometimes switched to English. We detected these cases using a language detection library and regenerated outputs with a temperature of 0.5 until at least 80\% of the output was identified as German. In cases where Llama 2 generated more than three items, we only kept the first three.

Appendix \ref{app:examples-same-text} contains examples of human-written and generated items for one of the texts.

\subsection{Human Evaluation}

We recruited six annotators for the human evaluation. All were university students or recent graduates and native German speakers. Considering that the texts and items in our dataset are targeted at CEFR level B1, it is safe to assume that the annotators can respond correctly to answerable items. The annotators took part on a voluntary basis and did not receive monetary compensation. The total workload was between 30 minutes and two hours per person.

We collected three types of annotations: (1) item responses without seeing the text, (2) item responses while seeing the text, and (3) item quality ratings.

Every annotator annotated all ten texts. For each text, the annotation involved two stages. In the \textbf{guessing stage}, the three items from \emph{one} generator (human, Llama 2, or GPT-4) were presented, and the annotator was asked to guess for each answer option whether it is correct or incorrect. The reason for only showing the items from a single generator is that the items from different generators would often contain very similar questions, but with different answer options (see Appendix \ref{app:examples-same-text}). This meant that the answer to an item from one generator were sometimes guessable based on the set of answer options from another generator. In the \textbf{comprehension stage}, the text and the items from \emph{all} generators were shown. Annotators were asked to respond to the items again and additionally rate the quality of each item on a scale from 1 (unusable) to 5 (perfect). The following criteria for quality were listed, but annotators were free in how they weighted the criteria:

\begin{itemize}
  \item The item refers to the content of the text.
  \item The item is comprehensible and grammatically correct.
  \item The item is unambiguously answerable.
  \item The item is answerable without additional world knowledge.
  \item The item is only answerable after reading the text (not through world knowledge alone).
\end{itemize}

We randomized the order of the texts, items, and answer options for each annotator. Screenshots of the evaluation interface are shown in Appendix \ref{app:user-interface}.

\subsection{Automatic (LLM-Based) Evaluation}
\label{sec:setup:automatic-evaluation}

We used zero-shot prompting to elicit item responses from Llama 2 and GPT-4 in two settings. In the \textbf{comprehension setting}, each prompt contained the text, the stem of a single item, and a single answer option, and the models were instructed to respond with a binary (true/false) label. In the \textbf{guessing setting}, the prompt did not include the text (refer to Appendix \ref{app:prompts} for the full prompts). Both settings used a sampling temperature of 0.

Note that this procedure is different from the human evaluation in that only a single answer option is shown at a time. The main reason for this is to simplify parsing the LLM output. Particularly with Llama 2, showing all answer options and prompting the model to list all correct answer options in a consistent way was not feasible.

While GPT-4 consistently produced responses in the requested format, Llama 2 frequently responded with wordy disclaimers (e.g., ``Without seeing the text, it is difficult to say \ldots''). To bypass this behavior for Llama 2, we compared the predicted probabilities (i.e., softmaxed output scores) for the first generated token to determine which label was more likely.

Llama 2 showed a strong bias towards positive responses. We therefore considered the response to be positive if $P(\textit{true}) / (P(\textit{true}) + P(\textit{false})) \geq \tau$. We optimized the threshold $\tau$ to maximize response accuracy in each setting separately on an additional 50 texts from the same dataset. The resulting thresholds were $\tau_\text{with text} = 0.9952$ and $\tau_\text{without text} = 0.9849$. No such optimization was done for GPT-4.

The code for the automatic evaluation is available on GitHub\footnote{\url{https://github.com/saeub/item-evaluation}}.

\section{Results}

\subsection{Text Informativity}
\label{sec:results:text-informativity}

\begin{figure}
  \centering
  \includegraphics[width=\columnwidth]{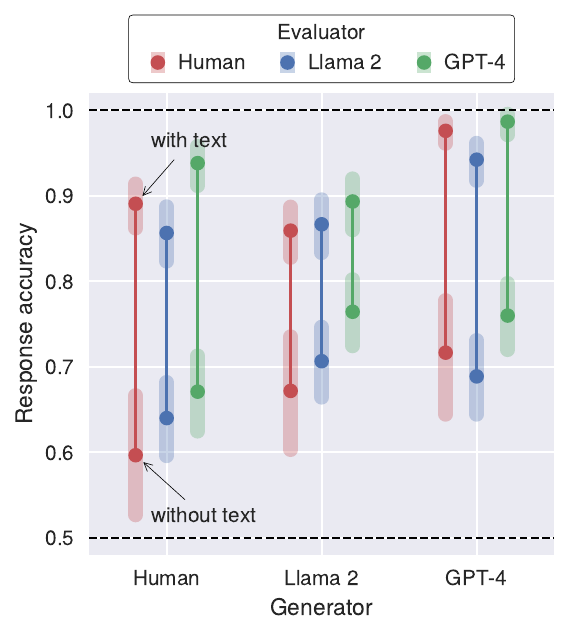}
  \caption{Mean human and LLM response accuracies on human-written and LLM-generated items. The distance between the two points corresponds to text informativity. Accuracies are on the level of answer options, therefore random guessing is at 0.5. For human evaluators, means are based on 10 texts and around 185 responses without text and around 546 responses with text. For LLM evaluators, means are based on 50 texts and around 451 responses in both settings. Error bars are bootstrapped 95\% confidence intervals.}
  \label{fig:ag-accuracy}
\end{figure}

Figure \ref{fig:ag-accuracy} shows the guessability and answerability estimates for the items of the three generators (human, Llama 2, and GPT-4) according to the three evaluators (humans, Llama 2, and GPT-4). For easier comparability, the text informativity metrics are also reported in Table \ref{tab:informativity}.

\begin{table}
  \centering
  \input{tables/informativity.tex}
  \caption{Text informativity (\textuparrow) for all combinations of generators and evaluators. The best text informativity estimates per evaluator are marked in bold. Numbers in brackets are bootstrapped 95\% confidence intervals.}
  \label{tab:informativity}
\end{table}

The three evaluators agree on several observations: human-written items have the lowest guessability, items generated by GPT-4 have the highest answerability, and items generated by Llama 2 have the lowest text informativity.

Overall, GPT-4 as an evaluator outperformed humans in terms of response accuracy both when guessing and when seeing the text. However, since text informativity is the difference between the accuracies in both settings, this difference in performance has little effect on text informativity, as evidenced by the similar values in Table \ref{tab:informativity} between human and GPT-4 evaluators. Llama 2 appears to be less reliable in this respect.

\subsection{Quality Ratings}

\begin{figure*}
  \centering
  \begin{subfigure}[t]{0.45\textwidth}
    \centering
    \includegraphics[scale=0.9]{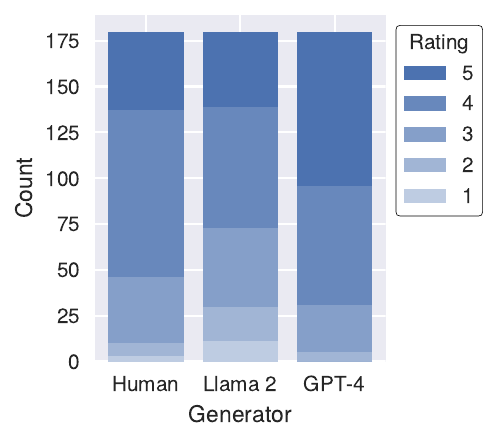}
    \caption{Rating distributions for different item generators. On average, items generated by GPT-4 received the highest ratings, Llama 2 the lowest.}
    \label{fig:ratings}
  \end{subfigure}
  \hfill
  \begin{subfigure}[t]{0.45\textwidth}
    \centering
    \includegraphics[scale=0.9]{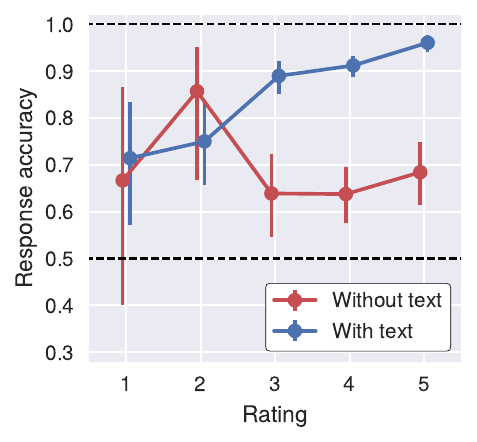}
    \caption{Mean human response accuracies with and without text grouped by item rating (irrespective of generator). Items rated higher tend to have better answerability. Error bars are bootstrapped 95\% confidence intervals.}
    \label{fig:ratings-accuracies}
  \end{subfigure}
  \caption{Distributions of human quality ratings and their relation to human response accuracy. A rating of 1 means \emph{unusable}, 5 means \emph{perfect}.}
\end{figure*}

The distribution of human quality ratings is shown in Figure \ref{fig:ratings}. Across all generators, more than half of the ratings were \emph{good} (4) or \emph{perfect} (5), indicating that most items were of acceptable quality. Items generated by Llama 2 were rated the worst on average, with 11/180 \emph{unusable} (1) ratings. Surprisingly, GPT-4 received considerably more \emph{perfect} ratings (84/180) than human and Llama 2 items.

Comparing the ratings to the response accuracies in Figure \ref{fig:ratings-accuracies} reveals that highly rated items tended to have higher answerability, while there is no clear relationship between ratings and guessability. This suggests that the annotators prioritized answerability over guessability in their ratings and explains the higher ratings for items generated by GPT-4, which tended to be both highly answerable and easily guessable (see Figure \ref{fig:ag-accuracy}).

\subsection{Inter-Annotator Agreement}

To quantify how human-like the responses by the LLM evaluators are, we measured the agreement between the two models and the group of human annotators. To achieve this, we calculated pairwise inter-annotator agreements (IAAs) using Cohen's $\kappa$ between the binary responses from the LLM and each of the humans, for both Llama 2 and GPT-4. We then compared the mean of this pairwise model-human IAA to the mean human-human IAAs. If the model-human IAA is similar to the human-human IAAs, this indicates that the model's response behavior is similar to that of the human annotators.

The results in Table \ref{tab:iaa} show that GPT-4 provided the most human-like responses and even exceeded the average human-human IAA in both settings. IAA between Llama 2 and humans was lower, but still within the range of human-human IAAs.

\begin{table}
  \centering
  \input{tables/iaa.tex}
  \caption{Mean Cohen's $\kappa$ between responses by evaluators and (other) humans with and without text across items (irrespective of generator). The IAAs in the setting with text are based on 93 binary responses (10 texts, 30 items). The values in the setting without text are less reliable because each human only annotated a third of all items in this setting. The mean pairwise agreement between GPT-4 and humans (0.724) is larger than the average agreement between the six humans (0.683).}
  \label{tab:iaa}
\end{table}

\subsection{Qualitative Analysis}

To provide a tangible explanation for why some items are more guessable or less answerable than others, we conducted a qualitative analysis of generated and human-written items that were either guessed correctly without the text or answered incorrectly with the text by a majority of human annotators. We describe the most common phenomena here and refer to Appendix \ref{app:examples-guessable-unanswerable} for specific examples.

The main reason why items are highly guessable is that they ask about real-world concepts or events that are widely known even without reading the text. This is especially common in our dataset because the texts are news articles about current events. Items that are difficult to guess tend to involve questions about the text itself rather than the events described in it. Examples of such questions are ``What is the text about?'' or ``What does the text say about \ldots?'' where all answer options may be plausible, but not all are true given the text. For most texts, there is at least one question of this type among the human-written items in our dataset, while GPT-4 and Llama 2 tend to generate fewer of these questions.

The explanations for items not being perfectly answerable are more diverse. We found three common features of unanswerable items, listed here in descending order of frequency:

\begin{enumerate}
  \item \textbf{Wrong label:} The item has an incorrect \emph{true}/\emph{false} label for some answer options. This occurred most frequently with Llama 2, and especially when none of the generated answer options are correct, but the model still produced the \emph{true} label for one of them.
  \item \textbf{Unclear answer options:} The item is phrased in a way that leaves room for interpretation. In particular, some answer options paraphrase information from a text such that not all annotators may agree that they still bear the same meaning.
  \item \textbf{Insufficient evidence:} The text does not provide the necessary evidence to decide conclusively whether an answer option is correct. In many of these cases, answering correctly requires additional world knowledge.
\end{enumerate}

\section{Discussion}

\subsection{LLMs for Item Generation}

One of the aims of this paper was to evaluate LLMs for zero-shot generation of MCRC items in German. Given the lack of data, zero- and few-shot learning are the most promising techniques for this language, and our results strongly suggest that state-of-the-art instruction-tuned LLMs are capable of generating items of acceptable quality. In particular, items generated by GPT-4 are close to the human-written items in our dataset in terms of text informativity. Llama 2 also produced noteworthy results, considering that only 0.17\% of the pre-training data is German \citep{Touvron2023}, this is still an impressive result. Using more multilingual or German-centric LLMs could further improve this performance.

A common problem with both models was that they produced easily guessable items, as our evaluations showed. Guessability can be measured in a straightforward manner with human or LLM annotators, and this feedback could be used to improve AIG performance in future work, e.g., through reinforcement learning from human or model feedback \citep{Ouyang2022,Bai2022}. Previously, \citet{Yuan2017} and \citet{Klein2019} have used similar approaches to improve the answerability of generated items.

\subsection{Evaluation Protocol}

We presented a simple method and a metric for evaluating reading comprehension items. There are several advantages to our method in comparison to previous work. Compared to ratings, this method is more objective and human-centric. It is also more meaningful and interpretable than similarity-based metrics like BLEU, and it does not rely on references. Another advantage is that the same protocol can be used for human and automatic evaluation.

One of the most important limitations is that text informativity only considers two aspects of item quality, i.e., answerability and guessability. Although these are some of the most difficult and critical criteria to meet, there are other aspects that can lead to low item quality \citep{Jones2020}. For example, our approach cannot detect items where the correct answer options use the same wording as in the text, meaning that no comprehension is required for a correct response. Some of these cases can easily be detected, e.g., using string matching. Characteristics such as grammaticality and difficulty would also have to be addressed separately. We leave these for future work.

Another limitation is that the protocol relies on highly proficient test takers, while the test items in our dataset are targeted at language learners. This is by design, as the goal is to measure the items' answerability given that the text was fully understood, but it still means that the response behavior in the human evaluation is not representative of the target user group.

\subsection{LLMs for Item Evaluation}

The evaluation protocol we presented measures guessability and answerability by item responses from human annotators with and without showing them the text. By replacing the humans with LLMs, we are making two assumptions:

\begin{enumerate}
  \item The LLMs have similar world knowledge to humans, resulting in similar guessability estimates.
  \item The LLMs have similar reading comprehension abilities to humans, resulting in similar answerability estimates.
\end{enumerate}

Based on the results presented in Section \ref{sec:results:text-informativity}, using GPT-4 leads to an over-estimation of both guessability and answerability. In contrast to previous work focusing only on answerability \citep{Yuan2017,Klein2019,Shuai2021,Rathod2022,Raina2022a,Uto2023}, using text informativity as a metric normalizes this difference to some degree. The high IAA between GPT-4 and human annotators also suggest that using GPT-4 as an evaluator is a viable option. In contrast, results from Llama 2 were less consistent with humans, both at the dataset level and the response level. Moreover, Llama 2 only yielded usable results after optimizing the classification threshold on additional data as described in Section \ref{sec:setup:automatic-evaluation} (meaning that the responses were not technically zero-shot in this case). However, depending on the use case, it may still be a good open-source option for evaluation.

Compared to previous work \citep{Berzak2020,Liusie2023,Raina2023}, using LLMs for estimating answerability and guessability has several advantages: since we use zero-shot generation, no training is required. This is particularly convenient for languages such as German, where no large MCRC datasets exist. Zero-shot generation also prevents overfitting on dataset-specific features that would go unnoticed by human test takers (compare \citet{Berzak2020}, where a fine-tuned RoBERTa classifier consistently outperformed human test takers at guessing the correct answer).

A limitation of our approach is that a single LLM is unable to capture human label variation. On the one hand, this means that we cannot model how strongly different human annotators will agree on their responses to a specific item, which can be useful for evaluation \citep{Plank2022}. On the other hand, it means that evaluating the quality of a single item is not feasible, which is why we only reported text informativity at the level of an entire dataset. Possible solutions to this problem include using multiple models \citep{Lalor2019,Byrd2022} or prompt variation \citep{PortilloWightman2023} to determine uncertainty.

\section{Conclusion and Future Work}

The overarching goal of this paper was to explore the potential of LLMs for generating and evaluating MCRC items. To this end, we introduced a new evaluation protocol and metric, text informativity, and demonstrated its applicability for both human and automatic evaluation. We used this protocol to evaluate two state-of-the-art LLMs for zero-shot item generation based on a dataset of German texts and MCRC items from online language courses. Our results show that both GPT-4 and Llama 2 are capable of generating items of acceptable quality, but GPT-4 clearly outperforms in terms of text informativity and human quality ratings. We also found that using GPT-4 for automatic evaluation is a viable option, while Llama 2 is less reliable.

These insights have significant implications: they show that zero-shot learning can make automatic item generation and evaluation feasible in languages where MCRC resources are scarce. Our evaluation protocol also addresses the lack of automatic evaluation metrics for the task. In a more general sense, using LLMs to generate reading comprehension items \emph{and} to predict how humans will respond to these items is a promising approach -- not only for language assessment in education, but also for comprehensibility evaluation in text simplification and readability assessment.

Future work could focus on improving item generation, e.g., by using text informativity as a reward for reinforcement learning, or improving item evaluation, e.g., by making LLM responses more human-like and reflective of individual variability and uncertainty.

\section{Acknowledgements}

We acklowledge support from the Department of Computational Linguistics at the University of Zurich for providing computational resources for this research. We also thank the annotators for their time and effort in evaluating the items. Finally, we thank the anonymous reviewers for their valuable feedback. This work was partially funded by the Swiss Innovation Agency (Innosuisse) Flagship IICT (PFFS-21-47).

\section{Bibliographical References}

\bibliographystyle{lrec-coling2024-natbib}
\bibliography{references.bib}

\appendix
\onecolumn
\raggedbottom

\clearpage
\section{Prompts}
\label{app:prompts}

The instructions used for generating items and responses for the automatic evaluation are specified in Tables \ref{tab:generation-prompt} and \ref{tab:evaluation-prompt}. For both models, the instructions were provided as the first user message, and no system instructions were specified.

\begin{table}[h]
    \centering
    \begin{tabular}{p{7cm}p{7cm}}
        \toprule
        German & English \\
        \midrule
        Text: \newline
        \tcbox{[$T$]} \newline
        \newline
        Schreibe 3 Multiple-Choice-Verständnisfragen zum Text oben, in deutscher Sprache. Jede Frage soll 3 Antwortmöglichkeiten haben. Schreibe hinter jede Antwort in Klammern, ob sie richtig oder falsch ist. Zwischen 0 und 3 Antworten können richtig sein. Die falschen Antworten sollten plausibel sein, wenn man den Text nicht gelesen hat.
        &
        Text: \newline
        \tcbox{[$T$]} \newline
        \newline
        Write 3 multiple-choice comprehension questions about the text above, in German language. Each question should have 3 answer options. After each answer, write whether it is correct or incorrect in parentheses. Between 0 and 3 answers can be correct. The incorrect answers should be plausible, not having read the text.
        \\
        \bottomrule
    \end{tabular}
    \caption{The German prompt template for item generation and a translation into English. In the text $T$, headings and paragraphs were separated by a newline character.}
    \label{tab:generation-prompt}
\end{table}

\begin{table}[h]
    \centering
    \begin{tabular}{lp{6.5cm}p{6.5cm}}
        \toprule
        & German & English \\
        \midrule
        \rotatebox[origin=l]{-90}{\textbf{With text}}
        &
        Text: \newline
        \tcbox{[$T$]} \newline
        \newline
        Frage: \tcbox{[$q$]} \newline
        Antwort: \tcbox{[$a$]} \newline
        \newline
        Gemäß dem Text oben, ist diese Antwort richtig (R) oder falsch (F)? Gib nur den Buchstaben R oder F an.
        &
        Text: \newline
        \tcbox{[$T$]} \newline
        \newline
        Question: \tcbox{[$q$]} \newline
        Answer: \tcbox{[$a$]} \newline
        \newline
        Based on the text above, is this answer correct (C) or incorrect (I)? Indicate only the letter C or I.
        \\
        \midrule
        \rotatebox[origin=l]{-90}{\textbf{Without text}}
        &
        Die folgende Frage und Antwort stammen aus einer Multiple-Choice-Verständnisaufgabe zu einem unbekannten Text. \newline
        \newline
        Frage: \tcbox{[$q$]} \newline
        Antwort: \tcbox{[$a$]} \newline
        \newline
        Ohne den Text zu kennen, nur basierend auf Allgemeinwissen, ist es plausibler, dass die Antwort richtig (R) oder falsch (F) ist? Gib nur den Buchstaben R oder F an.
        &
        The following question and answer are from a multiple-choice comprehension task about an unknown text. \newline
        \newline
        Question: \tcbox{[$q$]} \newline
        Answer: \tcbox{[$a$]} \newline
        \newline
        Without knowing the text, only based on general knowledge, is this answer more likely to be correct (C) or incorrect (I)? Indicate only the letter C or I.
        \\
        \bottomrule
    \end{tabular}
    \caption{The German prompt templates for item evaluation and a translation into English. In the text $T$, headings and paragraphs were separated by a newline character.}
    \label{tab:evaluation-prompt}
\end{table}

\clearpage
\section{Examples of Human-Written and Generated Items for the Same Text}
\label{app:examples-same-text}

The following sections show all human-written and generated items for one of the texts in the test set. The check marks (\cmark) and crosses (\xmark) indicate whether the answer option is correct or incorrect (according to the author/generator).

The corresponding lesson on the DW website (including the German text) can be found at \url{https://learngerman.dw.com/de/l-46996604}. The text is about Yemen's national football team, who had qualified for the Asia Cup in the United Arab Emirates, but faced challenges preparing for the championship due to political tensions.

\subsection{Human-Written Items}

\begin{tabular}{p{7.5cm}p{7.5cm}}
  German & English \\
  \midrule
  \raggedright
  Der Text handelt vor allem von ...
  \begin{mcanswers}
    \item[\xmark] Fußballfans im Jemen und wie sie versuchen, die Vereine in ihren Orten zu unterstützen.
    \item[\xmark] einer großen nationalen Sportveranstaltung, die im Jemen stattfinden sollte.
    \item[\cmark] den Vorbereitungen der jemenitischen Nationalmannschaft auf eine wichtige Meisterschaft.
  \end{mcanswers}
  &
  The text is mainly about ...
  \begin{mcanswers}
    \item[\xmark] Football fans in Yemen and how they try to support the clubs in their towns.
    \item[\xmark] a major national sports event that was supposed to take place in Yemen.
    \item[\cmark] the preparations of the Yemeni national team for an important championship.
  \end{mcanswers}
  \\
  Was sagt Bashir Sinan?
  \begin{mcanswers}
    \item[\cmark] Wenn Fußball gespielt wird, vergessen die Jemeniten den Krieg in ihrem Land.
    \item[\cmark] Das jemenitische Fußballteam leidet unter den politischen Problemen im Land.
    \item[\xmark] Für die Fußballer ist das nächste Ziel, die Asienmeisterschaften zu gewinnen.
  \end{mcanswers}
  &
  What does Bashir Sinan say?
  \begin{mcanswers}
    \item[\cmark] When playing football, Yemenis forget about the war in their country.
    \item[\cmark] The Yemeni football team suffers from the political problems in the country.
    \item[\xmark] The next goal for the football players is to win the Asia Cup.
  \end{mcanswers}
  \\
  Was ist richtig?
  \begin{mcanswers}
    \item[\cmark] Viele Jemeniten fliehen vor dem Bürgerkrieg in ihrem Land.
    \item[\cmark] Die jemenitische Mannschaft will, dass es Friedensgespräche gibt.
    \item[\xmark] Katar und Saudi-Arabien kämpfen im Krieg gemeinsam gegen den Jemen.
  \end{mcanswers}
  &
  What is correct?
  \begin{mcanswers}
    \item[\cmark] Many Yemenis are fleeing the civil war in their country.
    \item[\cmark] The Yemeni team wants peace talks to take place.
    \item[\xmark] Qatar and Saudi Arabia are fighting together in the war against Yemen.
  \end{mcanswers}
  \\
\end{tabular}

\subsection{Items Generated by Llama 2}

\begin{tabular}{p{7.5cm}p{7.5cm}}
  German & English \\
  \midrule
  \raggedright
  Warum dürfen keine Fußballspiele mehr in Jemen stattfinden?
  \begin{mcanswers}
    \item[\cmark] Weil das Land zu unsicher ist
    \item[\xmark] Weil es ein Bürgerkrieg gibt
    \item[\xmark] Weil die Bevölkerung nicht interessiert ist
  \end{mcanswers}
  &
  Why are no more football games allowed in Yemen?
  \begin{mcanswers}
    \item[\cmark] Because the country is too unsafe
    \item[\xmark] Because there is a civil war
    \item[\xmark] Because the population is not interested
  \end{mcanswers}
  \\
  Wie haben die jemenitischen Fußballspieler sich für die Asienmeisterschaften qualifiziert?
  \begin{mcanswers}
    \item[\cmark] Sie haben in verschiedenen arabischen Ländern trainiert
    \item[\xmark] Sie haben ein wildcard [\emph{sic}] erhalten
    \item[\xmark] Sie haben sich nicht qualifiziert
  \end{mcanswers}
  &
  How did the Yemeni football players qualify for the Asia Cup?
  \begin{mcanswers}
    \item[\cmark] They trained in various Arab countries
    \item[\xmark] They received a wildcard
    \item[\xmark] They did not qualify
  \end{mcanswers}
  \\
  Was ist die Botschaft der jemenitischen Fußballmannschaft an die Bevölkerung und die Politik?
  \begin{mcanswers}
    \item[\cmark] Lassen wir die Gewehre schweigen und einen vernünftigen Dialog und Frieden an ihre Stelle treten
    \item[\xmark] Lassen wir die Gewalt weitergehen und uns nichts ausreden
    \item[\xmark] Lassen wir uns auf die politischen Spannungen einigen und die Fußballmeisterschaft boykottieren
  \end{mcanswers}
  &
  What is the message of the Yemeni football team to the population and the politicians?
  \begin{mcanswers}
    \item[\cmark] Let's silence the guns and have a reasonable dialogue and peace take their place
    \item[\xmark] Let's let the violence continue and not talk ourselves out of anything
    \item[\xmark] Let's agree on the political tensions and boycott the football championship
  \end{mcanswers}
  \\
\end{tabular}

\subsection{Items Generated by GPT-4}

\begin{tabular}{p{7.5cm}p{7.5cm}}
  German & English \\
  \midrule
  \raggedright
  Warum dürfen im Jemen keine Fußballspiele mehr stattfinden?
  \begin{mcanswers}
    \item[\cmark] Wegen des Bürgerkriegs und der damit verbundenen Unsicherheit.
    \item[\xmark] Weil die Fußballstadien zerstört wurden.
    \item[\xmark] Weil die jemenitischen Fußballspieler alle das Land verlassen haben.
  \end{mcanswers}
  &
  Why are no more football games allowed in Yemen?
  \begin{mcanswers}
    \item[\cmark] Because of the civil war and the associated insecurity.
    \item[\xmark] Because the football stadiums were destroyed.
    \item[\xmark] Because all Yemeni football players have left the country.
  \end{mcanswers}
  \\
  Wo fand das Training für die Asienmeisterschaften statt?
  \begin{mcanswers}
    \item[\cmark] In verschiedenen arabischen Ländern, wie Saudi-Arabien und Katar.
    \item[\cmark] In Malaysia.
    \item[\xmark] Im Jemen.
  \end{mcanswers}
  &
  Where did the training for the Asia Cup take place?
  \begin{mcanswers}
    \item[\cmark] In various Arab countries, such as Saudi Arabia and Qatar.
    \item[\cmark] In Malaysia.
    \item[\xmark] In Yemen.
  \end{mcanswers}
  \\
  Was ist die Botschaft der jemenitischen Fußballmannschaft an die Bevölkerung und die Politik?
  \begin{mcanswers}
    \item[\xmark] Sie sollten mehr Geld in den Fußball investieren.
    \item[\cmark] Sie sollten die Gewehre schweigen lassen und einen vernünftigen Dialog und Frieden an ihre Stelle treten.
    \item[\xmark] Sie sollten die Fußballspieler besser unterstützen.
  \end{mcanswers}
  &
  What is the message of the Yemeni football team to the population and the politicians?
  \begin{mcanswers}
    \item[\xmark] They should invest more money in football.
    \item[\cmark] They should silence the guns and have a reasonable dialogue and peace take their place.
    \item[\xmark] They should support the football players better.
  \end{mcanswers}
  \\
\end{tabular}

\clearpage
\section{User Interface for Human Evaluation}
\label{app:user-interface}

\begin{figure}[h]
    \centering
    \includegraphics[width=0.6\textwidth]{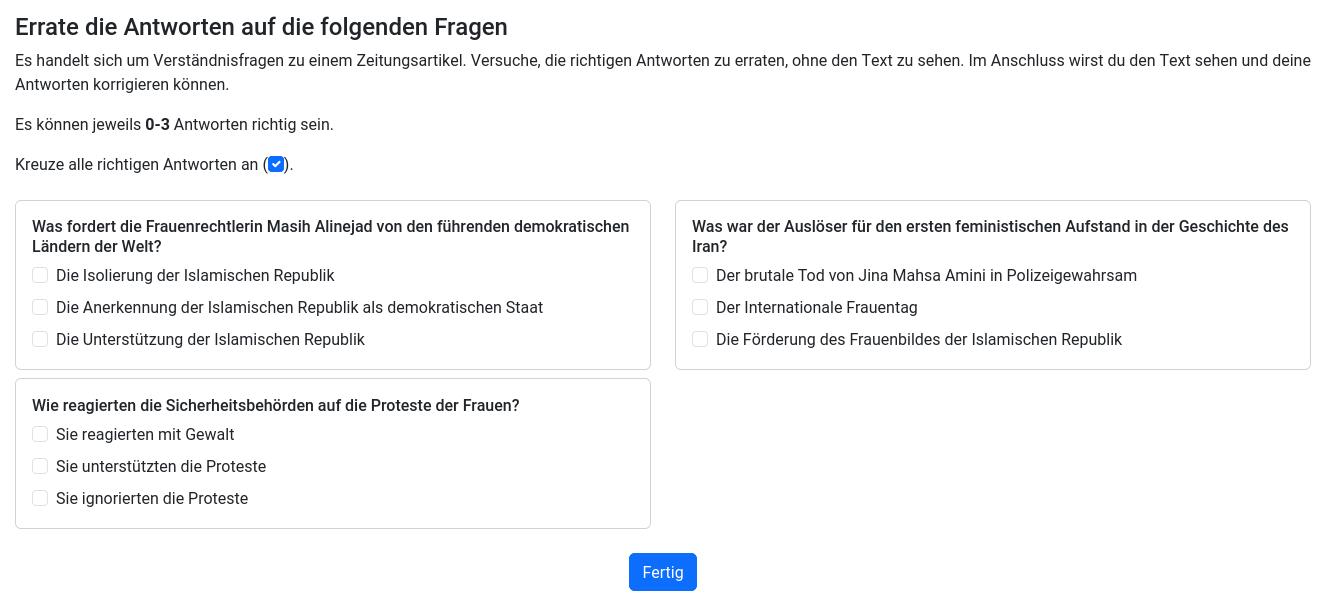}
    \caption{Screenshot of the user interface for the human evaluation, without text.}
\end{figure}

\begin{figure}[h]
    \centering
    \includegraphics[width=0.6\textwidth]{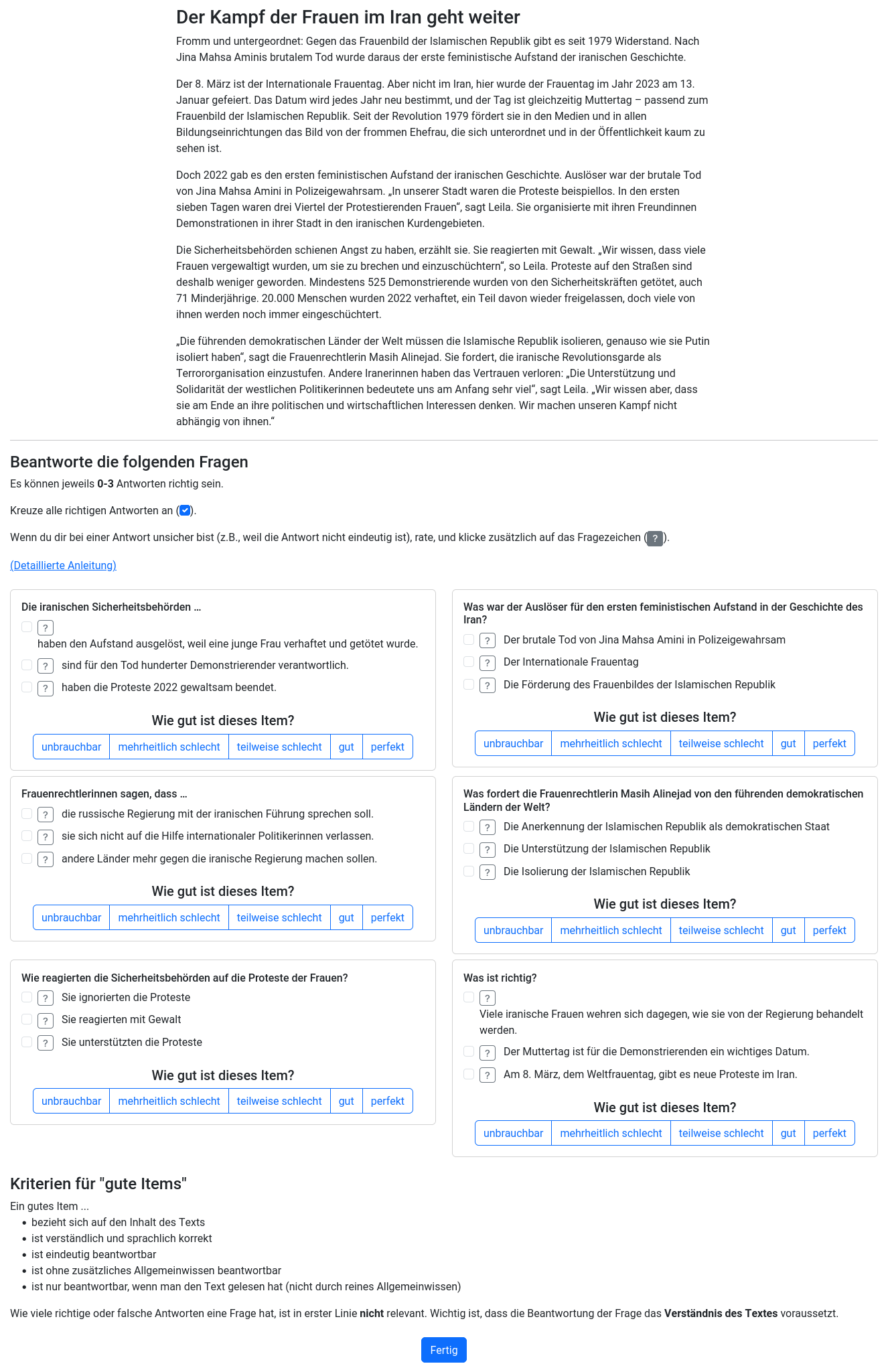}
    \caption{Screenshot of the user interface for the human evaluation, with text and quality ratings.}
\end{figure}

\clearpage
\section{Examples of Guessable and Unanswerable Items}
\label{app:examples-guessable-unanswerable}

\subsection{Guessable Items}

\begin{tabular}{p{2.5cm}p{6cm}p{6cm}}
  & German & English \\
  \midrule
  \raggedright
  \textbf{Item} \newline (human-written)
  &
  Die Saturnalien ...
  \begin{mcanswers}
    \item[\xmark] wurden an den längsten Tagen im Jahr gefeiert.
    \item[\cmark] waren ein Fest, bei dem unfreie Menschen mit den Herrschern die Rollen wechselten.
    \item[\cmark] fanden unter anderem in der römischen Stadt Köln statt.
  \end{mcanswers}
  &
  The Saturnalia ...
  \begin{mcanswers}
    \item[\xmark] were celebrated on the longest days of the year.
    \item[\cmark] were a festival where unfree people switched roles with the rulers.
    \item[\cmark] took place in the Roman city of Cologne, among other places.
  \end{mcanswers}
  \\
  \midrule
  \raggedright
  \textbf{Item} \newline (generated by GPT-4)
  &
  Was bedeutet der Name „Karneval“ aus dem Lateinischen übersetzt?
  \begin{mcanswers}
    \item[\xmark] „Fest der Freude“
    \item[\cmark] „Fleisch, leb wohl“
    \item[\xmark] „Tanz der Narren“
  \end{mcanswers}
  &
  What does the name ``Carnival'' mean when translated from Latin?
  \begin{mcanswers}
    \item[\xmark] ``Festival of joy''
    \item[\cmark] „Flesh, farewell“
    \item[\xmark] „Dance of the fools“
  \end{mcanswers}
  \\
\end{tabular}

\subsection{Unanswerable Items}

\subsubsection{Wrong label}

\begin{tabular}{p{2.5cm}p{6cm}p{6cm}}
  & German & English \\
  \midrule
  \textbf{Text excerpt}
  &
  {[...]} \textit{Die Musikwissenschaftlerin Marina Schwarz meint dazu: „Das ist Teil der immer noch patriarchalischen Gesellschaft, in der wir leben.“ Offenbar finden auch viele Frauen, die in dieser Gesellschaft aufgewachsen sind, solche Texte normal.} {[...]}
  &
  {[...]} \textit{Musicologist Marina Schwarz says: ``This is part of the still patriarchal society in which we live.'' Apparently, many women who have grown up in this society also find such lyrics normal.} {[...]}
  \\
  \midrule
  \raggedright
  \textbf{Item} \newline (generated by Llama 2)
  &
  Was ist laut Text Marina Schwarz' Meinung zu sexistischen Texten im Schlager?
  \begin{mcanswers}
    \item[\cmark] Sie findet sie inakzeptabel. {[should be \xmark]}
    \item[\xmark] Sie findet sie normal, weil es Teil der patriarchalischen Gesellschaft ist.
    \item[\xmark] Sie findet sie nicht sexistisch, sondern nur humorvoll.
  \end{mcanswers}
  &
  According to the text, what is Marina Schwarz' opinion on sexist lyrics in Schlager?
  \begin{mcanswers}
    \item[\cmark] She finds them unacceptable. {[should be \xmark]}
    \item[\xmark] She finds them normal, because it is part of the patriarchal society.
    \item[\xmark] She does not find them sexist, just humorous.
  \end{mcanswers}
  \\
\end{tabular}

\subsubsection{Unclear answer options}

\begin{tabular}{p{2.5cm}p{6cm}p{6cm}}
  & German & English \\
  \midrule
  \textbf{Text excerpt}
  &
  {[...]} \textit{Für viele Deutsche zählt beim Kiosk eher die Atmosphäre -- besonders in der warmen Jahreszeit.} {[...]}
  &
  {[...]} \textit{For many Germans, the atmosphere is more important at the kiosk -- especially in the warm season.} {[...]}
  \\
  \midrule
  \raggedright
  \textbf{Item} \newline (human-written)
  &
  Viele Menschen kaufen Alkohol am Kiosk, weil ...
  \begin{mcanswers}
    \item[\cmark] er dort billiger ist als in Bars und Kneipen.
    \item[\cmark] sie die schöne Stimmung vor Ort mögen. {[unclear if \cmark or \xmark]}
    \item[\cmark] sie auf dem Weg zu einer Party etwas trinken möchten.
  \end{mcanswers}
  &
  Many people buy alcohol at the kiosk because ...
  \begin{mcanswers}
    \item[\cmark] it is cheaper there than in bars and pubs.
    \item[\cmark] they like the nice atmosphere on site. {[unclear if \cmark or \xmark]}
    \item[\cmark] they want to drink something on the way to a party.
  \end{mcanswers}
  \\
\end{tabular}

\subsubsection{Insufficient evidence}

\begin{tabular}{p{2.5cm}p{6cm}p{6cm}}
  & German & English \\
  \midrule
  \textbf{Text excerpt}
  &
  {[...]} \textit{Besonders im Rheinland sind die Straßen voll mit kostümierten Menschen, die tanzen, singen und feiern} {[...]}
  &
  {[...]} \textit{Especially in the Rhineland, the streets are full of people in costumes who dance, sing and celebrate} {[...]}
  \\
  \midrule
  \raggedright
  \textbf{Item} \newline (generated by Llama 2)
  &
  Wo finden die meisten Karnevalsumzüge und -feiern statt?
  \begin{mcanswers}
    \item[\cmark] In Köln
    \item[\xmark] In Rom
    \item[\xmark] In Berlin
  \end{mcanswers}
  &
  Where do most carnival parades and celebrations take place?
  \begin{mcanswers}
    \item[\cmark] In Cologne
    \item[\xmark] In Rome
    \item[\xmark] In Berlin
  \end{mcanswers}
  \\
\end{tabular}

\end{document}

%% file: tables/informativity.tex
\begin{tabular}{rrccc}
\toprule
 & & \multicolumn{3}{c}{Evaluator} \\
 & & Human & Llama 2 & GPT-4 \\
\midrule
\multirow[c]{6}{*}{\hskip-0.2cm\rotatebox{90}{Generator}} & Human & \bfseries 0.294 & 0.216 & \bfseries 0.267 \\
 & & \scriptsize \raisebox{1ex}{[0.220, 0.367]} & \scriptsize \raisebox{1ex}{[0.161, 0.272]} & \scriptsize \raisebox{1ex}{[0.219, 0.316]} \\
 & Llama 2 & 0.187 & 0.160 & 0.129 \\
 & & \scriptsize \raisebox{1ex}{[0.115, 0.262]} & \scriptsize \raisebox{1ex}{[0.109, 0.213]} & \scriptsize \raisebox{1ex}{[0.082, 0.178]} \\
 & GPT-4 & 0.259 & \bfseries 0.253 & 0.227 \\
 & & \scriptsize \raisebox{1ex}{[0.193, 0.328]} & \scriptsize \raisebox{1ex}{[0.204, 0.302]} & \scriptsize \raisebox{1ex}{[0.187, 0.269]} \\[-1ex]
\bottomrule
\end{tabular}

%% file: tables/iaa.tex
\begin{tabular}{lrr}
\toprule
\multicolumn{3}{r}{Mean IAA with (other) humans} \\
Evaluator & without text & with text \\
\midrule
Human 1 & 0.185 & 0.712 \\
Human 2 & 0.015 & 0.679 \\
Human 3 & 0.000 & 0.677 \\
Human 4 & 0.400 & 0.669 \\
Human 5 & 0.000 & 0.634 \\
Human 6 & 0.216 & 0.729 \\
\midrule
Humans 1--6 (average) & 0.136 & 0.683 \\
Llama 2 & 0.051 & 0.651 \\
GPT-4 & \textbf{0.185} & \textbf{0.724} \\
\bottomrule
\end{tabular}